\documentclass[runningheads]{llncs}
\usepackage{graphicx}
\usepackage{amsmath,amssymb} 
\usepackage{color}
\usepackage[width=122mm,left=12mm,paperwidth=146mm,height=193mm,top=12mm,paperheight=217mm]{geometry}

\usepackage{nicefrac}       
\usepackage{microtype}      
\usepackage{multirow}
\usepackage{verbatim}
\usepackage{color}
\usepackage{float}
\usepackage{enumitem}
\usepackage{booktabs}
\usepackage{tabulary,multirow,overpic,xcolor}
\usepackage[caption=false]{subfig}

\usepackage[british,UKenglish,USenglish,english,american]{babel}
\usepackage[pagebackref=true,breaklinks=true,letterpaper=true,colorlinks,bookmarks=false]{hyperref}

\newcommand{\ve}[1]{\mathbf{#1}} 
\newcommand{\ma}[1]{\mathrm{#1}} 

\newcolumntype{x}[1]{>{\centering\arraybackslash}p{#1pt}}

\newcommand{\bd}[1]{\textbf{#1}}
\newcommand{\app}{\raise.17ex\hbox{$\scriptstyle\sim$}}

\def\x{$\times$}
\newcolumntype{x}[1]{>{\centering\arraybackslash}p{#1pt}}

\newlength\savewidth\newcommand\shline{\noalign{\global\savewidth\arrayrulewidth
  \global\arrayrulewidth 1pt}\hline\noalign{\global\arrayrulewidth\savewidth}}
\newcommand{\tablestyle}[2]{\setlength{\tabcolsep}{#1}\renewcommand{\arraystretch}{#2}\centering\footnotesize}

\makeatletter\renewcommand\paragraph{\@startsection{paragraph}{4}{\z@}
  {.5em \@plus1ex \@minus.2ex}{-.5em}{\normalfont\normalsize\bfseries}}\makeatother

\begin{document}

\pagestyle{headings}
\mainmatter

\title{Videos as Space-Time Region Graphs} 

\author{Xiaolong Wang, Abhinav Gupta}

\institute{Robotics Institute, Carnegie Mellon University}

\maketitle

\begin{figure}
\centering
\includegraphics[width=1\linewidth]{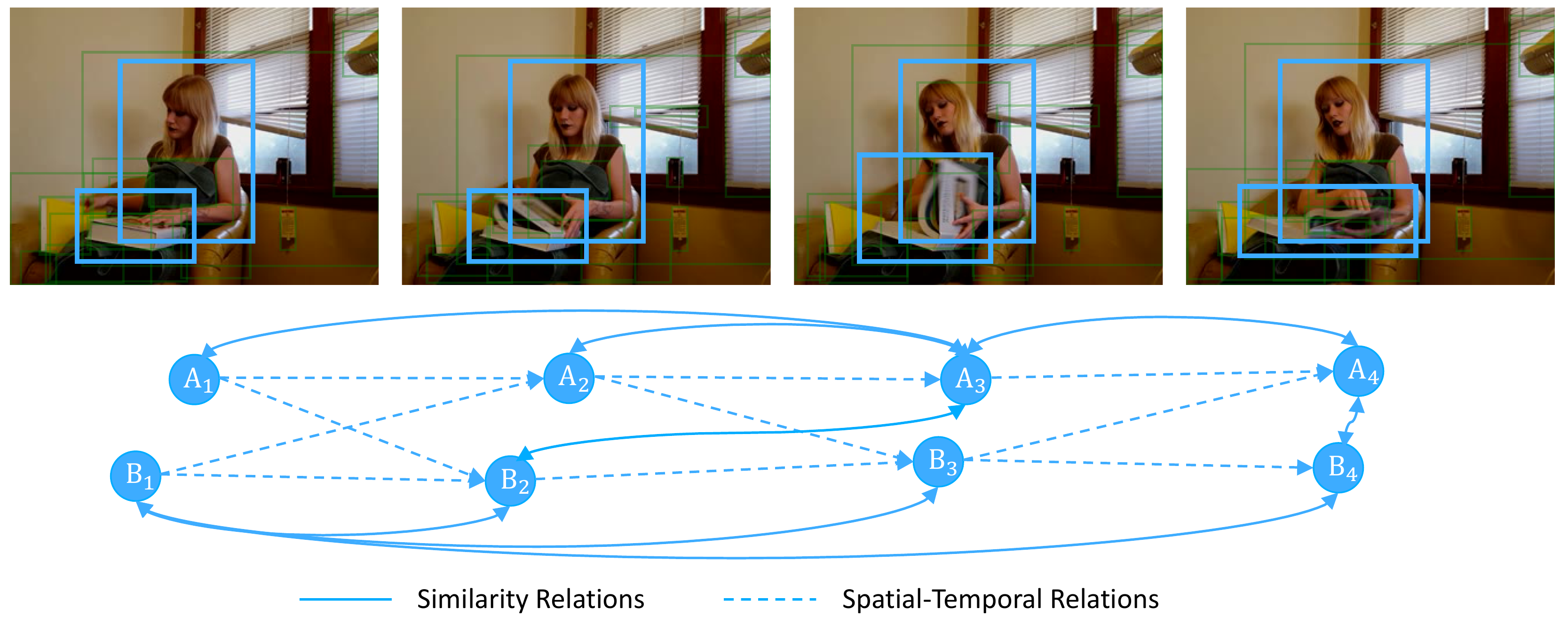}
\vspace{-0.1in}
\caption{How do you recognize simple actions such as opening book? We argue action understanding requires appearance modeling but also capturing temporal dynamics (how shape of book changes) and functional relationships. We propose to represent videos as space-time region graphs followed by graph convolutions for inference. }
\label{fig:teaser}
\vspace{-1em}
\end{figure}

\begin{abstract}

How do humans recognize the action ``opening a book''? We argue that there are two important cues: modeling temporal shape dynamics and modeling functional relationships between humans and objects. In this paper, we propose to represent videos as space-time region graphs which capture these two important cues. Our graph nodes are defined by the object region proposals from different frames in a long range video. These nodes are connected by two types of relations: (i) similarity relations capturing the long range dependencies between correlated objects and (ii) spatial-temporal relations capturing the interactions between nearby objects. We perform reasoning on this graph representation via Graph Convolutional Networks. We achieve state-of-the-art results on both Charades and Something-Something datasets. Especially for Charades, we obtain a huge $4.4\%$ gain when our model is applied in complex environments. 

\end{abstract}

\section{Introduction}

Consider a simple action such as ``opening a book'' as shown in Fig.~\ref{fig:teaser}. When we humans see the sequence of images, we can easily recognize the action category; yet our current vision systems (with hundreds of layers of 3D convolutions) struggle on this simple task. Why is that? What is missing in current video recognition frameworks?

Let's first take a closer look at the sequence shown in Fig.~\ref{fig:teaser}. How do humans recognize the action in the video corresponds to ``opening a book''? We argue that there are two key ingredients to solving this problem: First, the shape of the book and how it changes over time (i.e., the object state changes from closed to open)  is a crucial cue. Exploiting this cue requires temporally linking book regions across time and modeling actions as transformations. But just modeling temporal dynamics of objects is not sufficient. The state of objects change after interaction with human or other objects. Thus we also need to model human-object and object-object interactions as well for action recognition. 

However, our current deep learning approaches fail to capture these two key ingredients. For example, the state-of-the-art approaches based on two-stream ConvNets~\cite{Simonyan2014,WangXWQLTV16} are still learning to classify actions based on individual video frame or local motion vectors. Local motion clearly fails to model the dynamics of shape changes. To tackle this limitation, recent work has also focused on modeling long term temporal information with Recurrent Neural Networks~\cite{Yue-HeiNg2015,Donahue2015LRCN,li2017temporal,miech2017learnable} and 3D Convolutions~\cite{Tran2015,Carreira2017,Tran18,Xie17}. However, all these frameworks focus on the features extracted from the whole scenes and fail to capture long-range 
temporal dependencies (transformations) or region-based relationships. In fact, most of the actions are classified based on the background information instead of capturing the key objects (e.g., the book in ``opening a book'') as observed in~\cite{sigurdsson2017actions}. 

On the other hand, there have been several efforts to specifically model the human-object or object-object interactions~\cite{Gupta09PAMI,Yao10CVPR}. This direction have been recently revisited with ConvNets in an effort to improve object detection~\cite{yatskar2016,HuCVPR18,gkioxari2017interactnet}, visual relationship detection~\cite{Lu16ECCV} and action recognition~\cite{rstarcnn}, etc. However, the relationship reasoning is still performed in static images failing to capture temporal dynamics of these interactions. Thus, it is very hard for these approaches to capture the changes of object states over time as well as the causes and effects of these changes. 

In this paper, we propose to perform long-range temporal modeling of human-object and object-object relationships via a graph-based reasoning framework. Unlike existing approaches which focus on local motion vectors, our model takes in a long range video sequence (e.g., more than 100 frames or 5 seconds). We represent the input video as a {\bf space-time region graph} where each node in the graph represent region of interest in the video. Region nodes are connected by two types of edges: appearance-similarity and spatio-temporal proximity. Specifically, \textbf{(i) Similarity Relations}: regions which have similar appearance or semantically related are connected together. With similarity relations, we can model how the states of the same object change and the long range dependencies between any two objects in any frames.  \textbf{(ii) Spatial-Temporal Relations}: objects which overlap in space and close in time are connected together via these edges. With spatial-temporal relations, we can capture the interactions between nearby objects as well as the temporal ordering of object state changes. 

Given the graph representation, we perform reasoning on the graph and infer the action by applying the Graph Convolution Networks  (GCNs)~\cite{kipf2017semi}. We conduct our experiments in the challenging Charades~\cite{Sigurdsson2016} and 20BN-Something-Something~\cite{20bnsthsth} datasets. Both datasets are extremely challenging as the actions cannot be easily inferred by the background of the scene and the 2D appearance of the objects or humans. Our model shows significant improvements over state-of-the-art results of action recognition. Especially in the Charades dataset, we obtain $4.4\%$ boost.  

Our contributions include: (a) A novel graph representation with variant relationships between different objects in a long range video; (b) A graph convolutional network model for reasoning with multiple relation edges; (c) state-of-the-art performance with a significant gain in action recognition in complex environments.

\vspace{-0.1in}
\section{Related Work}
\vspace{-0.1in}

\textbf{Video Understanding Models.} Spatio-temporal reasoning is one of the core research areas in the field of video understanding and action recognition. However, most of the early work has focused on using spatio-temporal appearance features. For example, a large effort has been spent on manually designing the video features ~\cite{STIP05,Wang2013a,HOG3D,MBH06,HOF,Corso12,YangGreg11,Zhuowen13,Peng14,Lan15}. Some of the hand-designed features such as the Improved Dense Trajectory (IDT)~\cite{Wang2013a} are still widely applied and show very competitive results in different video related tasks. However, instead of designing hand-crafted features, recent researches have focused towards learning deep representations from the video data~\cite{Karpathy14,Simonyan2014,TDD15,Taylor10,Le11,Zhou2017,WangXWQLTV16,Wang_Transformation}. One of the most popular model is the two-Stream ConvNets~\cite{Simonyan2014} where temporal information is model by a network with 10 optical flow frames as inputs ($<1$ second). To better model longer-term information, a lot of work has been focused on using Recurrent Neural Networks (RNNs)~\cite{Yue-HeiNg2015,Donahue2015LRCN,Srivastava15,Sun15,Wu15,li2017temporal,gan2016you,Bian2017,pan2016hierarchical} and 3D ConvNets~\cite{3DCNN,Tran15,Carreira2017,Tran18,Xie2017,Qiu17,Feichtenhofer2016}. However, these frameworks focus on extracting features from the whole scenes and can hardly model the relationships between different object instances in space and time. 

\textbf{Visual Relationships.} Reasoning about the pairwise relationships has been proven to be very helpful in a variety of computer vision tasks~\cite{Gupta09PAMI,Yao10CVPR,Yao12Det,Kumar10,russell2006using}. For example, object detection in cluttered scenes can be significantly improved by modeling the human-object interactions~\cite{Yao10CVPR}. Recently, the visual relationships have been widely applied together with deep networks in the area of visual question answering~\cite{Santoro2017}, object recognition~\cite{yatskar2016,HuCVPR18,gkioxari2017interactnet} and  intuitive physics~\cite{Battaglia2016,Watters2017}. In the case of action recognition, a lot of effort has been made on modeling pairwise human-object and object-object relationships~\cite{Ma2018,deepparts,ni2016progressively}. However, the interaction reasoning framework in these efforts focus on static images and the temporal information is usually modeled by a RNN on image level features. Thus, these approaches still cannot capture how a certain object state changes or rather transformations over time. 

An attempt at modeling pairwise relations in space and time has been recently made in the Non-local Neural Networks~\cite{xiaolongwang2017nonlocal}. However, the Non-local operator is applied in every pixel in the feature space (from low layers to higher layers), while our reasoning is based on a graph with object level features. Moreover, the non-local operator does not process any temporal ordering information, while this is explicit modeled in our spatial-temporal relations. 

\textbf{Graphical Models.} The long range relationships in images and videos are usually captured by graphical models. One popular direction is using the Conditional Random Fields (CRF)~\cite{Lafferty2001,Kraehenbuehl2011}. In the context of deep learning, especially for semantic segmentation, the CRF model is often applied on the outputs of the ConvNets by performing mean-field inference~\cite{Chen2014,Zheng2015,Chandra2017,Schwing2015,krahenbuhl2011efficient,Harley2017}. Instead of using mean-field inference, variant simpler feedforward graph based neural network have been proposed recently~\cite{Liu2017,Scarselli2009,kipf2017semi,YujiaGSNN2016,Marino17,stgcn2018aaai}. In this paper, we apply the Graph Convolutional Networks (GCNs)~\cite{kipf2017semi} which was originally proposed for applications in Natural Language Processing. Our GCN is built by stacking multiple layers of graph convolutions with similarity relations and spatial-temporal relations. The outputs of the GCNs are updated features for each object node, which can be used to perform classification. 

Our work is also related to video recognition with object cues~\cite{Alayrac17,Wu16CVPR,Heilbron17} and object graph models~\cite{Brendel11,ChenCVPR2012,Jain16,YuanyuanICCV2017}. For example, Structural-RNN~\cite{Jain16} is proposed to model the spatial-temporal relations between objects (adjacent in time) for video recognition tasks. Different from these works, our space-time graph representation encodes not only local relations but also long range dependencies between any pairs of objects across space and time. By using graph convolutions with long range relations, it enables efficient message passing between starting states and ending states of the objects. This global graph reasoning framework provides significant boost over the state-of-the-art.

\vspace{-0.15in}
\section{Overview}
\vspace{-0.1in}
\begin{figure}[t]
\centering
\includegraphics[width=1\linewidth]{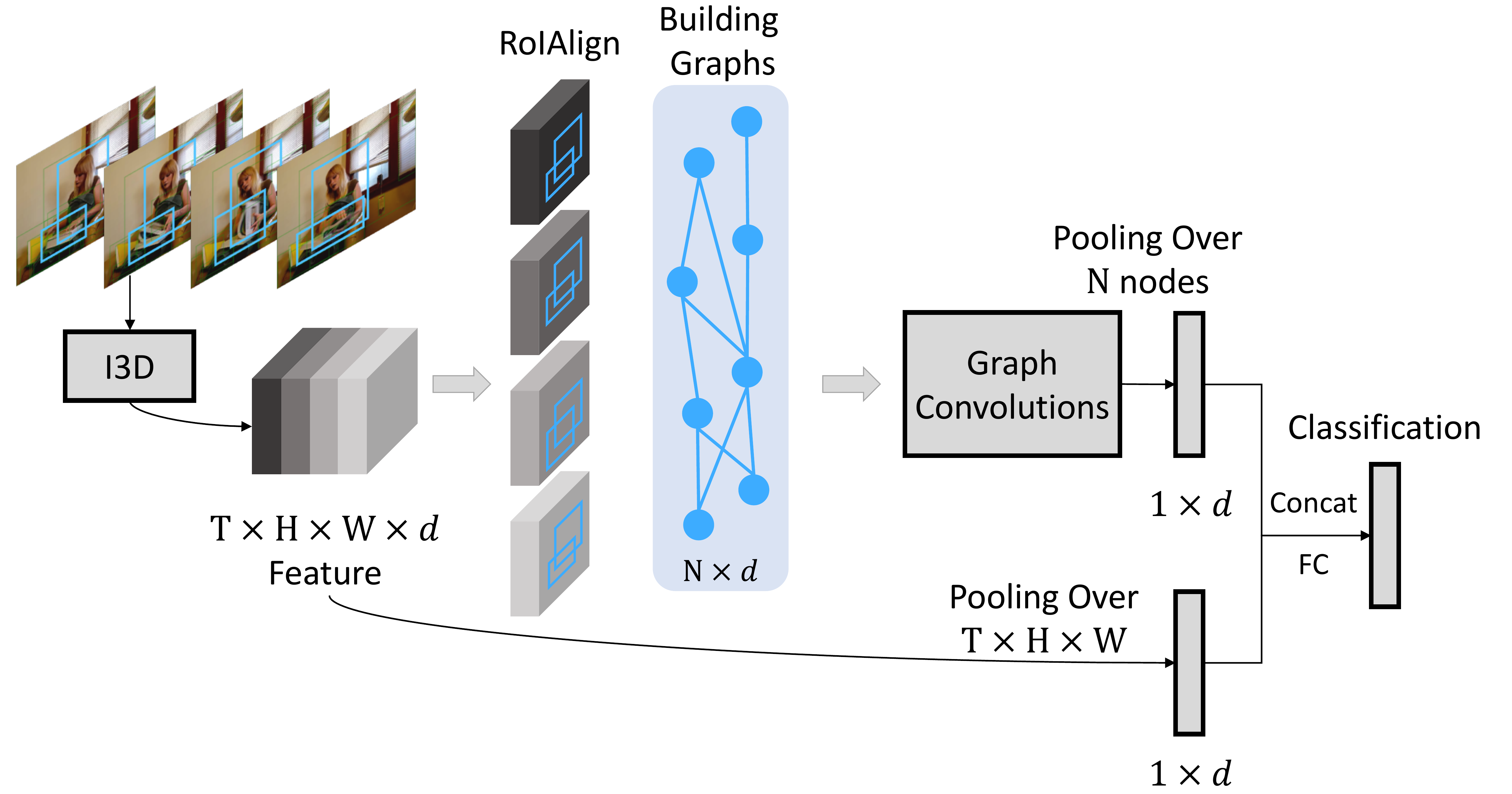}
\vspace{-2.1em}
\caption{Model Overview. Our model uses 3D convolutions to extract visual features followed by RoIAlign extracting $d$-dimension feature for each object proposal. These features are provided as inputs to the Graph Convolutional Network which performs information propagation based on spatiotemporal edges. Finally, a $d$-dimension feature is extracted and appended to another $d$-dimension video feature to perform classification.}
\label{fig:overview}
\vspace{-1em}
\end{figure}

Our goal is to represent the video as a graph of objects and perform reasoning on the graph for action recognition. The overview of our model is visualized in Figure~\ref{fig:overview}. Our model takes inputs as a long clip of video frames (more than 5 seconds) and forward them to a 3D Convolutional Neural Network~\cite{Carreira2017,xiaolongwang2017nonlocal}. The output of this 3D ConvNet is a feature map with the dimensions $\ma{T} \times \ma{H} \times \ma{W} \times d$, where $\ma{T}$ represents the temporal dimension, $\ma{H} \times \ma{W}$ represents the spatial dimensions and $d$ represents the channel number. 

Besides extracting the video features, we also apply a Region Proposal Network (RPN)~\cite{Ren2015} to extract the object bounding boxes (We have not visualized the RPN in Figure~\ref{fig:overview} for simplicity). Given the bounding boxes for each of the $\ma{T}$ feature frames, we apply  RoIAlign~\cite{Girshick2015,He2017} to extract the features for each bounding box.  Note that the RoIAlign is applied on each feature frame independently. The feature vector for each object has $d$ dimensions (first aligned to $7 \times 7 \times d$ and then maxpooled to $1 \times 1 \times  d$). We denote the object number as $\ma{N}$, thus the feature dimension is $\ma{N} \times d$ after RoIAlign.

We now construct a graph which contains $\ma{N}$ nodes corresponding to $\ma{N}$ object proposals aggregated over $\ma{T}$ frames. There are mainly two types of relations in the graph: similarity relations and spatial-temporal relations. For simplicity, we decompose this big graph into two sub-graphs with the same nodes but two different relations: the similarity graph and the spatial-temporal graph.  

With the graph representations, we apply the Graph Convolutional Networks (GCNs) to perform reasoning. The output for the GCNs are in the same dimension as the input features which is $\ma{N} \times d$. We perform average pooling over all the  object nodes to obtain a $d$-dimension feature. Besides the GCN features, we also perform average pooling on the whole video representation ($\ma{T} \times \ma{H} \times \ma{W} \times d$) to obtain the same $d$-dimension feature as a global feature. These two features are then concatenated together for video level classification.   

We will introduce the details of each component in the following sections. We introduce the process of feature extraction and graph representations in Section 4 and the Graph Convolutional Networks (GCNs) in Section 5.

\vspace{-0.1in}
\section{Graph Representations in Videos}
\vspace{-0.1in}

In this section, we will first introduce the feature extraction process for our model with 3D ConvNets and then describe the construction of the similarity graph as well as the spatial-temporal graph.

\newcommand{\blockb}[3]{\multirow{3}{*}{\(\left[\begin{array}{c}\text{3\x1\x1, #2}\\[-.1em] \text{1\x3\x3, #2}\\[-.1em] \text{1\x1\x1, #1}\end{array}\right]\)\x#3}}
\begin{table}[t]
\vspace{-0.1in}
\footnotesize
\centering
\resizebox{0.63\columnwidth}{!}{
\tablestyle{6pt}{1.08}
\begin{tabular}{c|c|c}
\multicolumn{2}{c|}{layer} & output size \\
\shline
conv$_1$ & \multicolumn{1}{c|}{5\x7\x7, 64, stride 1, 2, 2} & 32\x112\x112 \\
\hline
pool$_1$  & \multicolumn{1}{c|}{1\x3\x3 max, stride 1, 2, 2} & 32\x56\x56 \\
\hline
\multirow{3}{*}{res$_2$} & \blockb{256}{64}{3} & \multirow{3}{*}{32\x56\x56} \\
  &  & \\
  &  & \\
\hline
pool$_2$  & \multicolumn{1}{c|}{3\x1\x1 max, stride 2, 1, 1} & 16\x56\x56 \\
\hline
\multirow{3}{*}{res$_3$} & \blockb{512}{128}{4} & \multirow{3}{*}{16\x28\x28} \\
  &  & \\
  &  & \\
\hline
\multirow{3}{*}{res$_4$} & \blockb{1024}{256}{6} & \multirow{3}{*}{16\x14\x14}  \\
  &  & \\
  &  & \\
\hline
\multirow{3}{*}{res$_5$} & \blockb{2048}{512}{3} & \multirow{3}{*}{16\x14\x14} \\
  &  & \\
  &  & \\
\hline
\multicolumn{2}{c|}{global average pool, fc} & 1\x1\x1  \\
\end{tabular}}
\vspace{.5em}
\caption{Our baseline ResNet-50 I3D model. We use T\x H\x W to represent the dimensions of filter kernels and 3D output feature maps. For filter kernels, we also have number of channels following T\x H\x W. The input is in  32\x224\x224 dimensions and the residual blocks are shown in brackets.
}
\vspace{-0.3in}
\label{tab:arch}
\end{table}

\vspace{-0.1in}
\subsection{Video Representation}
\vspace{-0.03in}

\textbf{Video Backbone Model.} Given a long clip of video (around $5$ seconds), we sample $32$ video frames from it with the same temporal duration between every two frames. We extract the features on these frames via a 3D ConvNet. Table~\ref{tab:arch} shows our backbone model based on the ResNet-50 architecture, which are motivated by the model architecture mentioned in~\cite{xiaolongwang2017nonlocal}. The model takes input as $32$ video frames with $224 \times 224$ dimensions and the output of the last convolutional layer is a $16 \times 14 \times 14$ feature map (i.e., 16 frames in the temporal dimension and $14 \times 14$ in the  spatial dimension). The baseline method in this paper adopts the same architecture, and the classification is simply performed by using a global average pooling on the final convolutional features and then following by a fully connected layer. 

This backbone model is called Inflated 3D ConvNet (I3D)~\cite{Feichtenhofer2016,Carreira2017,xiaolongwang2017nonlocal} as one can turn a 2D ConvNet into a 3D ConvNet by inflating the kernels during initialization. That is, a 3D kernel with $t \times k \times k$ dimensions can be inflated from a 2D $k \times k$ kernel by copying the weights $t$ times and rescaling by $1/t$. Please refer to ~\cite{Feichtenhofer2016,Carreira2017,xiaolongwang2017nonlocal} for more initialization details.

\textbf{Region Proposal Network.} We apply the Region Proposal Network (RPN) in~\cite{Ren2015,Detectron2018} to generate the object bounding boxes of interest on each video frame. More specifically, we use the RPN with ResNet-50 backbone and FPN~\cite{Lin2017}. The RPN is pre-trained with the MSCOCO object detection dataset~\cite{Lin2014} and there is no weight sharing between the RPN and our I3D video backbone model. Note that the bounding boxes extracted by the RPN are class-agnostic.

To extract object features on top of the last convolutional layer, we project the bounding boxes from the 16 input RGB frames (which are sampled from the 32 input frames for I3D, with the sampling rates of 1 frame every 2 frames) to the 16 output feature frames. Taking the video features and projected bounding boxes, we apply RoIAlign~\cite{He2017} to extract the feature for each object proposal. Note that RoIAlign is similar to  RoIPooling~\cite{Girshick2015} which crops and rescales the object features into the same dimensions.  In RoIAlign, each output frame is processed independently.  The RoIAlign generates a $7\times7 \times d$ output features for each object which is then max-pooled to $1 \times 1 \times d$ dimensions.

\vspace{-0.1in}
\subsection{Similarity Graph}
\vspace{-0.05in}

\begin{figure}[t]
\centering
\includegraphics[width=1\linewidth]{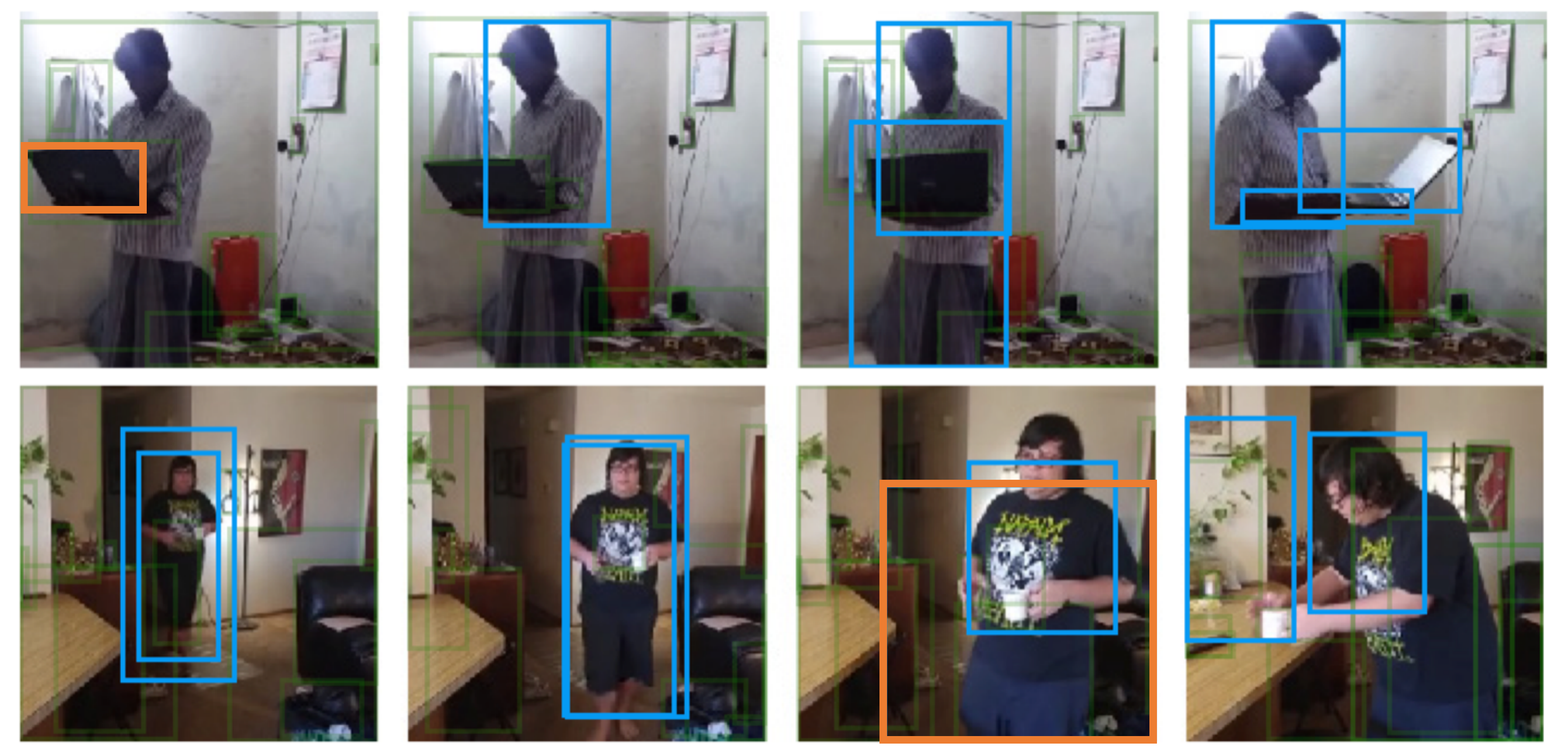}
\vspace{-1em}
\caption{Similarity Graph $\ve{G}^{sim}$. Above figure shows our similarity graph not only captures similarity in visual space but also correlations (similarity in functional space). The query box is shown in orange, the nearest neighbors are shown in blue. The transparent green boxes are the other unselected object proposals. }
\label{fig:simgraph}
\vspace{-1em}
\end{figure}

We measure the similarity between objects in the feature space to construct the similarity graph. In this graph, we connect pairs of semantically related objects together. More specifically, we will have a high confidence edge between two instances which are: (i) the same object in different states in different video frames or (ii) highly correlated for recognizing the actions. Note that the similarity edges are computed between any pairs of objects.

Formally, assuming we have the features for all the object proposals in the video as $\ve{X}=\{\ve{x}_1, \ve{x}_2, ..., \ve{x}_\ma{N}\}$, where $\ma{N}$ represents the number of object proposals and each  object proposal feature $\ve{x}_i$ is a $\emph{d}$ dimensional vector. The pairwise similarity or the affinity between every two proposals can be represented as,
\begin{equation}\label{eq:affinity}
F(\ve{x}_i, \ve{x}_j) = \phi(\ve{x}_i)^T\phi^{\prime}(\ve{x}_j),
\end{equation}
where $\phi$ and $\phi^{\prime}$ represents two different transformations of the original features. More specifically, we have $\phi(\ve{x}) =\ve{w}\ve{x}$ and $\phi^{\prime}(\ve{x}) = {\ve{w}^{\prime}\ve{x}}$.   The parameters $\ve{w}$ and $\ve{w}^{\prime}$ are both $d \times d$ dimensions weights which can be learned via back propagation.  By adding the transformation weights $\ve{w}$ and $\ve{w}^{\prime}$, it allows us to not only learn the correlations between different states of the same object instance across frame, but also the relations between different objects. We visualize the top nearest neighbors for the object proposals in Figure~\ref{fig:simgraph}. In the first example, we can see the nearest neighbors of the laptop  not only include the other laptop instances in other frames, but also the human who is operating it.

After computing the affinity matrix with Eq.~\ref{eq:affinity}, we perform normalization on each row of the matrix so that the sum of all the edge values connected to one proposal $i$ will be $1$. Motivated by the recent works~\cite{Vaswani2017,xiaolongwang2017nonlocal}, we adopt the softmax function for normalization as, 
\begin{equation}\label{eq:simg}
\ve{G}^{sim}_{ij} = \frac{\exp{F(\ve{x}_i, \ve{x}_j)}}{\sum_{j=1}^{\ma{N}} \exp{ F(\ve{x}_i, \ve{x}_j)} }. 
\end{equation}

The normalized $\ve{G}^{sim}$ is taken as the adjacency matrix representing the similarity graph. 

\vspace{-0.1in}
\subsection{Spatial-Temporal Graph}
\vspace{-0.05in}

\begin{figure}[t]
\centering
\includegraphics[width=1\linewidth]{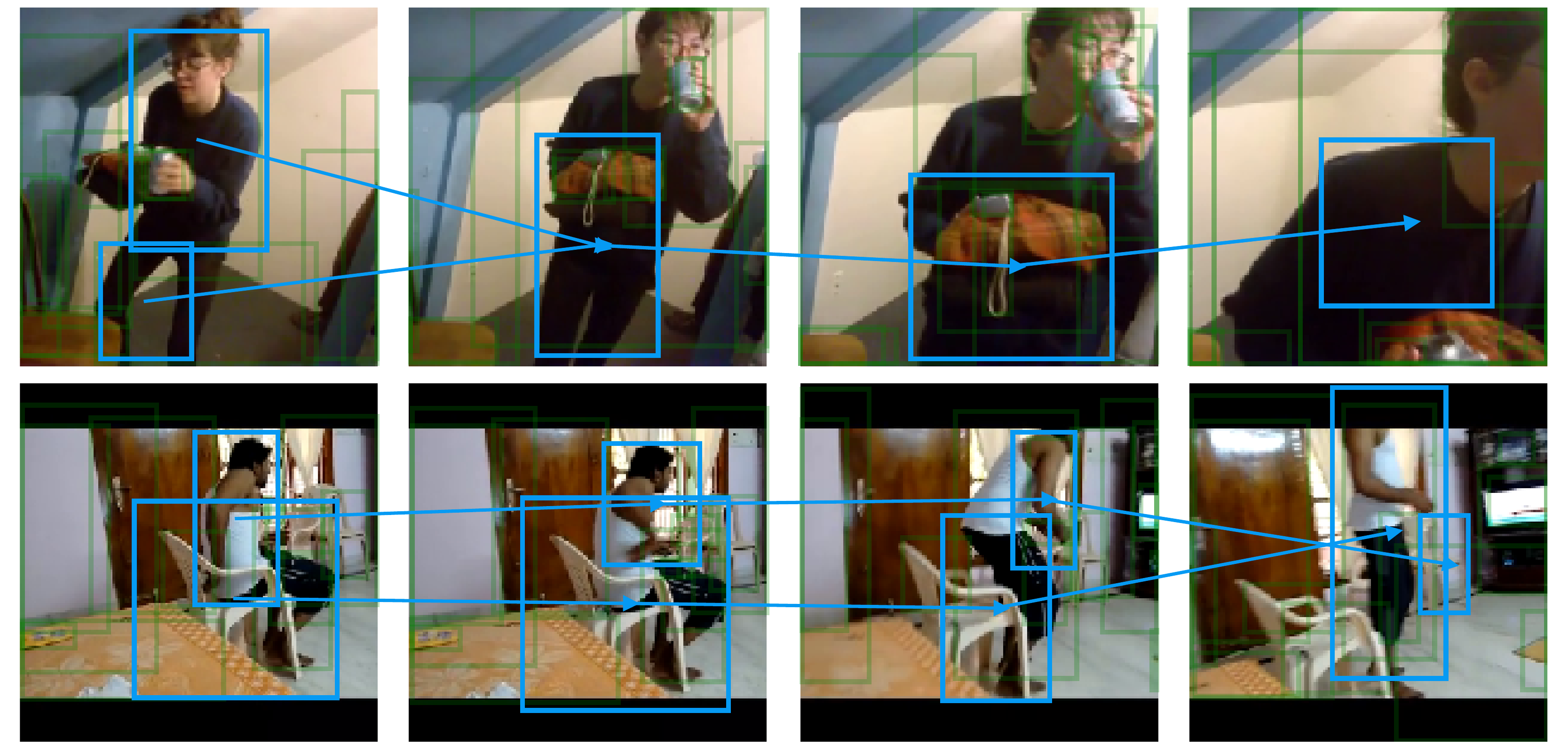}
\vspace{-1em}
\caption{Spatial-Temporal Graph $\ve{G}^{front}$. Highly overlapping object proposals across neighboring frames are linked by directed edge. We plot some example trajectories with blue boxes and the direction shows the arrow of time. }
\label{fig:timegraph}
\vspace{-1em}
\end{figure}

Although the similarity graph captures even the long term dependencies between any two object proposals, it does not capture the relative spatial relation between objects and the ordering of the state changes. To encode these spatial and temporal relations between objects, we propose to use spatial-temporal graphs, where objects in  nearby locations in space and time are connected together. 

Given a object proposal in frame $t$, we calculate the value of Intersection Over Unions (IoUs) between this object bounding box and all other object bounding boxes in frame $t+1$. We denote the IoU between object $i$ in frame $t$ and object $j$ in frame $t+1$ as $\sigma_{ij}$. If $\sigma_{ij}$ is larger than $0$, we will link object $i$ to object $j$ using a directed edge $i \rightarrow j$ with value $\sigma_{ij}$. After assigning the edge values, we normalize the graph so that the sum of the edge values connected to proposal $i$ will be $1$ by 
\begin{equation}\label{eq:timeg}
\ve{G}^{front}_{ij} = \frac{\sigma_{ij}}{\sum_{j=1}^{\ma{N}} \sigma_{ij}}, 
\end{equation}
where $\ve{G}^{front}$ is taken as the adjacency matrix for a spatial-temporal graph.  We visualize some of the object proposals and the trajectories in Figure~\ref{fig:timegraph}. 

Besides building the forward graph which connects objects from frame $t$ to frame $t+1$, we also construct a backward graph in a similar way which connect objects from frame $t+1$ to frame $t$. We denote the adjacency matrix of this backward graph as $\ve{G}^{back}$. Specifically, for the overlapping object $i$ in frame $t$ and object $j$ in frame $t+1$, we construct an edge $i \leftarrow j$ and assign the values to $\ve{G}^{back}_{ji}$ according to the IoU values. By building the spatial-temporal graphs in a bidirectional manner, we can obtain richer structure information and enlarge the number of propagation neighborhoods during graph convolutions.

\vspace{-0.1in}
\section{Convolutions on Graphs}
\vspace{-0.1in}

To perform reasoning on the graph, we apply the Graph Convolutional Networks (GCNs) proposed in~\cite{kipf2017semi}. Different from standard convolutions which operates on a local regular grid, the graph convolutions allow us to compute the response of a node based on its neighbors defined by the graph relations. Thus performing graph convolutions is equal to performing message passing inside the graphs. The outputs of the GCNs are updated features of each object node, which can be aggregated together for video classification. Formally, we can represent the one layer of graph convolutions as, 
\begin{equation}\label{eq:gcn}
\ve{Z} = \ve{G}\ve{X}\ve{W},
\end{equation}
where $\ve{G}$ represents one of the adjacency graph we have introduced ($\ve{G}^{sim}$, $\ve{G}^{front}$ or $\ve{G}^{back}$) with $\ma{N} \times \ma{N}$ dimensions,  $\ve{X}$ is the input features of the object nodes in the graph with $\ma{N} \times d$ dimensions, and $\ve{W}$ is the weight matrix of the layer with dimension $d \times d$ in our case. Thus the output of one graph convoltuional layer $\ve{Z}$ is still in $N \times d$ dimensions.  The graph convolution operation can be stacked into multiple layers. After each layer of graph convolutions, we apply two non-linear functions including the Layer Normalization~\cite{Ba2016} and then ReLU before the feature $\ve{Z}$ is forwarded to the next layer.

\textbf{Connecting GCN and Non-local Net.} We would also like to draw the connections between the Graph Convolutional Networks with the Similarity Graph ($\ve{G}^{sim}$) and the recent proposed Non-local Neural Networks~\cite{xiaolongwang2017nonlocal}. If we apply the non-local operation on the region proposals, it can be represented as, 
\begin{equation}\label{eq:nl1}
\ve{Y} = \ve{G}^{sim} g(\ve{X}),
\end{equation}
where $g$ is a function with a convolutional layer. The non-local block in~\cite{xiaolongwang2017nonlocal} can be further formulated as, 
\begin{equation}\label{eq:nl2}
\ve{Z} = \ve{Y} \ve{W} + \ve{X} = \ve{G}^{sim} g(\ve{X})\ve{W} + \ve{X},
\end{equation}
which is very similar to the graph convolution operation in Eq.~\ref{eq:gcn}. Inspired by this, we modify the graph convolution operations by adding a convolutional operator on the input $\ve{X}=g(\ve{X})$ for the first layer (note that $\ve{G}^{sim}$ is still computed based on the features before applying $g$). We also add a residual connection in every layer of GCN, which extends Eq.~\ref{eq:gcn} as, 
\begin{equation}\label{eq:gcn2}
\ve{Z} = \ve{G}\ve{X}\ve{W} + \ve{X}.
\end{equation}

\textbf{Combining Multiple Graphs.} To combine multiple graphs in GCNs, we can simply extend Eq.~\ref{eq:gcn2} as,
\begin{equation}\label{eq:gcnmulti}
\ve{Z} = \sum_{i} \ve{G}_i\ve{X}\ve{W}_i + \ve{X},
\end{equation}
where $\ve{G}_i$ indicates different types of graphs, and the weights for different graphs $\ve{W}_i$ are not shared. Note that in this way, each hidden layer of the GCN is updated though the relationships from different graphs. However, we find that the direct combination of 3 graphs ($\ve{G}^{sim}$, $\ve{G}^{front}$ and $\ve{G}^{back}$) with Eq.~\ref{eq:gcnmulti} actually hurts the performance compared to the situation with a single similarity graph. 

The reason is that our similarity graph $\ve{G}^{sim}$ contains learnable parameters (Eq.~\ref{eq:affinity}) and requires back propagation for updating, while the other two graphs do not require learning. Fusing these graphs together in every GCN layer increases the optimization difficulties. Thus we create two branches of graph convolutional networks, and only fuse the results from two GCNs in the end: one GCN adopts Eq.~\ref{eq:gcn} with $\ve{G}^{sim}$ and the other GCN adopts Eq.~\ref{eq:gcnmulti} with $\ve{G}^{front}$ and $\ve{G}^{back}$. These two branches of GCNs perform convolutions separately for $L$ layers and the final layer features are summed together, which is in $\ma{N} \times d$ dimensions. 

\textbf{Video Classification.} As illustrated in Figure~\ref{fig:overview}, the updated features after graph convolutions are forwarded to an average pooling layer, which calculates the mean of all the proposal features and leads to a $1 \times d$ dimension representation. Besides the GCN features, we also perform average pooling on the whole video level representation and obtain the another $1 \times d$ dimensions of global features. These two features are then concatenated together for video classification. The classification training loss is defined depending on the tasks (multi-label or single label classifications).

\vspace{-0.05in}
\section{Experiments}
We perform the experiments on two recent challenging datasets: Charades~\cite{Sigurdsson2016} and Something-Something~\cite{20bnsthsth}. We first introduce the implementation details of our approach and then the evaluation results on these datasets.

\vspace{-0.1in}
\subsection{Implementation Details} 
\vspace{-0.05in}

\paragraph{Training.} The training of our backbone models involves pre-training on 2 different datasets following~\cite{xiaolongwang2017nonlocal,Carreira2017}. The model is first pre-trained as a 2D ConvNet with the  ImageNet dataset~\cite{Russakovsky2015} and then inflated into a 3D ConvNet (i.e., I3D) as~\cite{Carreira2017}. We then fine-tuned the 3D ConvNet with the Kinetics action recognition dataset~\cite{Kay2017} following the same training scheme for longer sequences (around 5 second video) in~\cite{xiaolongwang2017nonlocal}. Given this initialization, we now introduce how to further fine-tune the network on our target datasets (e.g. Charades or Something-Something) as following. 

As specified in Table~\ref{tab:arch}, our network takes 32 video frames as inputs. These 32 video frames are sampled in the frame rate of 6fps, thus the temporal length of the video clip is around 5 seconds. The spatial dimensions for input is $224 \times 224$. Following~\cite{Simonyan2015}, the input frames are randomly cropped from a randomly scaled video whose shorter side is sampled in $[256, 320]$ dimensions. To reduce the number of GCN parameters, we add one more $1 \times 1 \times 1$ convolutional layer on top of the I3D  baseline model, which reduces the output channel number from $2048$ to $d=512$. Since both Charades and Something-Something dataset are in similar scales in number of video frames, we adopt the same learning rate schedule for both datasets. 

Our baseline I3D model is trained with a 4-GPU machine where each GPU has 2 video clips in a mini-batch. Thus the total batch size is 8 clips during training. Note that we freeze the parameters in all Batch Normalization (BN) layers during training. Our model is trained for 100K iterations in total, with learning rate $0.00125$ in the first 90K iterations and it is reduced by a factor of 10 during training the last 10K iterations. Dropout~\cite{Hinton2012} is applied on the last global pooling layer with a ratio of $0.3$. 

We set the layer number of our Graph Convolutional Network to $3$. The parameters of the convolutional operations $\phi, \phi^{\prime}$ and $g$ are initialized with Gaussian distribution having standard deviation of $0.01$. The parameters of kernels in graph convolutions $W$ are initialized  as zero inspired by~\cite{Goyal2017}. To train the GCN together with the I3D backbone, we propose to apply stage-wise training. We first finetune the I3D model as mentioned above, then we apply RoIAlign and GCN on top of the final convolutional features as shown in Figure~\ref{fig:overview}. We fix the I3D features and train the GCN with the same learning rate schedules as for training the backbone. Then we train the I3D and GCN together end-to-end for 30K more iterations with the reduced learning rate.

\emph{Task specific settings.} We apply different loss functions when training for Charades and Something-Something datasets. For Something-Something dataset, we can simply apply the softmax loss function. For Charades, we apply binary sigmoid loss, one for each action class, to handle the multi-label property. We also extract different numbers of object bounding boxes with RPN in two different datasets. As for Charades, the scenes are more cluttered and we extract 50 object proposals for each frame. However, for Something-Something, there is usually only one or two objects in the center of video frame and one hand is interacting with it. We find that extracting 10 object proposals each frame is enough for the Something-Something dataset.

\paragraph{Inference.} We perform fully-convolutional inference in space as~\cite{Simonyan2015,xiaolongwang2017nonlocal} during inference. Note that we rescale the shorter side of each video frame to $256$ while maintaining the aspect ratios. To perform inference on one whole video, we sample 10 clips for Charades and 2 clips for Something-Something according to the average video length in two different datasets. Results from multiple clips are aggregated together by Max-Pooling over the scores.

\vspace{-0.1in}
\subsection{Experiments on Charades}
\vspace{-0.05in}

In the Charades experiments, following the official split, we use the 8K training videos to train our model and perform testing on the 1.8K validation videos. The average video duration is around 30 seconds. There are 157 action classes and multiple actions can happen at the same time. 

\paragraph{How much each graph helps?} We first perform analysis on each component of our framework, with the backbone of ResNet-50 I3D, as illustrated in Table~\ref{tab:ablation:gcn}. We first show that the result of I3D baseline without any proposal extractions and graph convolutions is $31.8\%$ mAP on the validation set. 

One simple extension on this baseline is: obtain the region proposals with RPN, extract the features for each proposal and perform average pooling over them as an extra feature. We concatenate the video level feature and the proposal feature together for classification. However, we can only obtain $0.3\%$ boost with this approach. Thus, a naive aggregation of proposal features does not help much.  

We then perform evaluations by applying GCNs with the similarity graph and the  spatial-temporal graph individually. We observe that our GCN model with only spatial-temporal graph can obtain a boost of $2.4\%$ over the baseline model and achieve $34.2\%$. With the similarity graph, we can achieve a better performance of $35.0\%$. By combining two graphs together and train GCNs with multiple relations, our method achieves $36.2\%$ mAP which is a significant boost of $4.4\%$ over the baseline.

\begin{table*}[t]\centering
\subfloat[We perform ablation studies with GCN using the ResNet-50, I3D backbone. \label{tab:ablation:gcn}]{
\tablestyle{3pt}{1.05}
\begin{tabular}{l|x{55}}
\multicolumn{1}{c|}{model, R50, I3D}  & mAP  \\
\shline
baseline & 31.8   \\
\hline
Proposal+AvgPool & 32.1    \\
Spatial-Temporal GCN & 34.2  \\
Similarity GCN & 35.0  \\
Joint GCN & \bd{36.2} \\
\end{tabular}}\hspace{3mm}
\subfloat[We first compare our approach with Non-local Net and then combine Non-local Net with our model. \label{tab:ablation:gcnnl}]{
\tablestyle{2pt}{1.05}
\begin{tabular}{l|x{55}}
\multicolumn{1}{c|}{model, R50, I3D}  & mAP \\
\shline
baseline & 31.8 \\
\hline
Non-local  & 33.5  \\
Joint GCN & 36.2 \\
Non-local + Joint GCN & \bd{37.5}  \\
\multicolumn{2}{c}{~}\\
\end{tabular}}
\caption{\textbf{Ablations} on Charades. We show the mean Average Precision (mAP\%).}
\label{tab:ablations}
\vspace{-0.3in}
\end{table*}

\paragraph{Robustness to Proposal Numbers.} Besides studying on each sub-graph, we also analyze how the number of object proposals generated by the RPN affect our method. Note that the baseline achieves $31.8\%$ and our method achieves $36.2\%$ with extracting 50 object proposals per video frame. If we reduce the number of object proposals and extract 25 proposals per frame, the mAP of our method is $35.9\%$. If we increase and double the number of object proposals to 100 proposals per frame, the performance of our method is $36.1\%$ mAP. Thus our approach is actually very stable with the changes of RPN.

\paragraph{Model Complexity.} Given this large improvement in performance, the extra computation cost of the GCN over the baseline is actually very small. In the Charades dataset, our graph is defined based on 800 object nodes per video (with 16 output frames and 50 object proposals per frame). The computations of GCN with an 800-node graph is very small.  The FLOPs of the baseline I3D model is $153 \times 10^9$ and the total FLOPs of our model (I3D + Joint GCN) is $158 \times 10^9$. Thus there is only around $3\%$ increase in FLOPs. In fact, we barely observe training and inference time difference between baseline and our model.

\paragraph{Comparing to the Non-local Net.} One of the related work is the recent proposed Non-local Neural Networks~\cite{xiaolongwang2017nonlocal}, where they propose to perform non-local operations on different layers of feature maps for spatial-temporal reasoning. The comparisons between Non-local Nets and our approach is shown in Table~\ref{tab:ablation:gcnnl}. We can see that the Non-local operations gives $1.7\%$ improvements over the baseline and   our approach performs $2.7\%$ better than the Non-local Net. We also show that these two approaches are actually complementary to each other. By replacing the I3D backbone with Non-local Net, we have another $1.3\%$ boost, leading to $37.5\%$.  

\begin{figure}[t]
\centering
\includegraphics[width=0.85\linewidth]{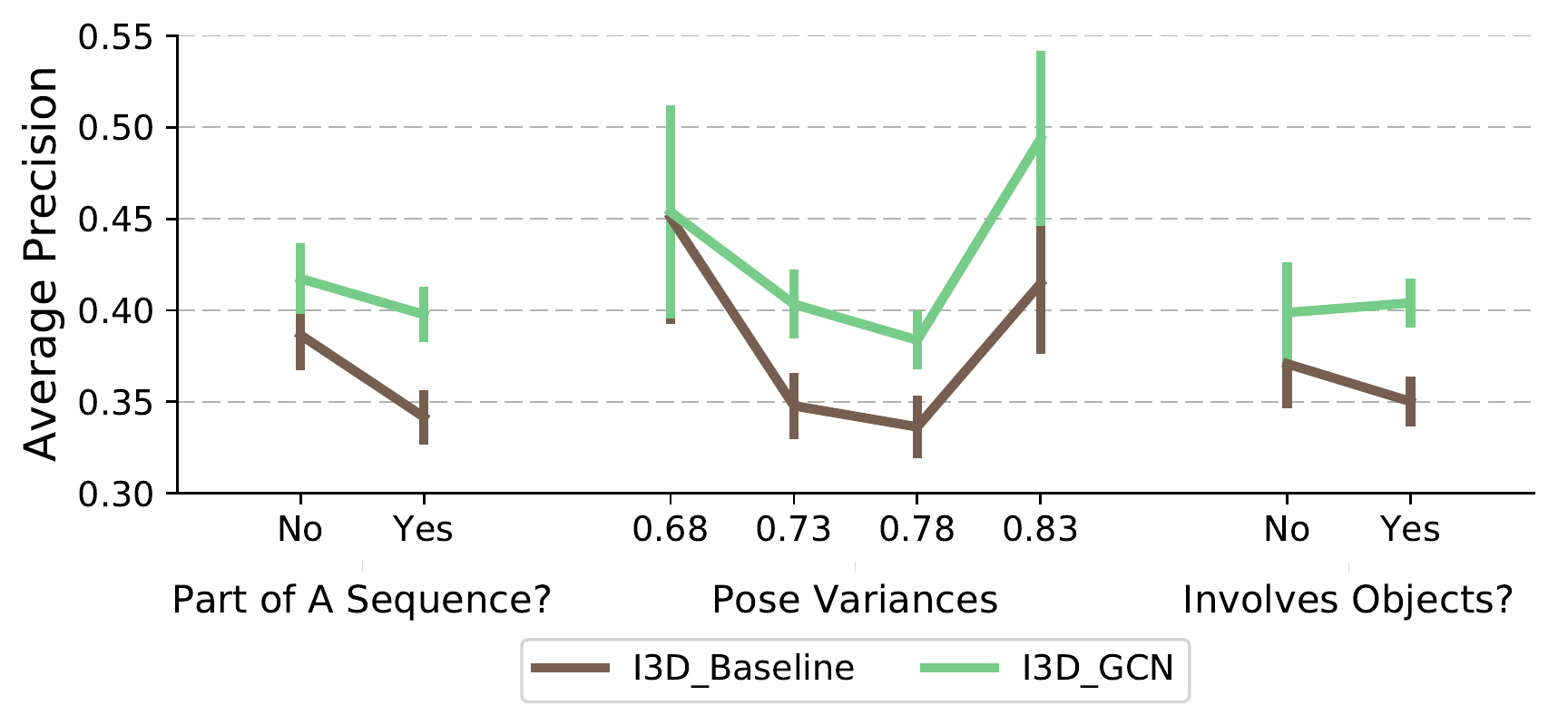}
\vspace{-1.2em}
\caption{Error Analysis. We compare our approach against baseline I3D approach across three different attributes. Our approach improves significantly when action is part of sequence, involves interaction with objects and has high pose variance. }
\label{fig:error}
\vspace{-1em}
\end{figure}

\paragraph{Error analysis}  
Given this significant improvements, we will also like to find out in what cases our methods improve over the baselines most. Following the attributes set up in~\cite{sigurdsson2017actions}, we show 3 different situations where our approach get more significant gains over the baselines in Figure~\ref{fig:error}. More specifically, for each video in Charades, besides the action class labels, it is also labeled with different attributes (e.g., whether the actions are happening in a sequence? Is the pose variant a lot though the actions? Is the action involving objects?).

\emph{Part of A Sequence?} This attribute specifies whether an action category is part of a sequence of actions. For example, ``holding a cup'' and then ``sitting down'' are usually in a sequence of actions, while ``running'' often happens in isolation. As shown in the left plots in Figure~\ref{fig:error}, the baseline I3D method fails dramatically when an action is part of a sequence of actions, while our approach is more stable.  If an action is not happening in isolation, we have actually more than $5\%$ gain over the baseline.

\emph{Pose Variances.} This attribute is computed by averaging the Procrustes distance~\cite{kendall1989survey} between any two poses in an action category. If the average distance is large, it means the poses change a lot in an action. As visualized in the middle plots in Figure~\ref{fig:error}, we can see that our approach has similar performance as the baseline when the pose variance is small. However, the performance of the baseline drops dramatically again as the variance of pose becomes larger (from $0.68$ to $0.73$) in the action, while the slope of our curve is much smaller. The performance of both approaches improve as the pose variability reaches $0.83$, where our approach has  around $8\% \sim 9\%$ boost over the baseline.

\emph{Involves Objects?} This attribute specifies whether an object is involved in the action. For example, ``drinking from a cup'' involves the object cup while ``running'' does not require interactions with objects. As shown in the right plots in Figure~\ref{fig:error}, we can see the baseline perform worse when the actions require interactions with objects.  Interestingly, our approach actually performs slightly better when objects are involved. 

As a short summary, our approach is better in modeling a long term sequence of actions and actions that require object interactions. Our approach is also more robust to pose changes and is able to utilize the motion from poses.

\begin{table}[t]
\centering
\small
\tablestyle{6pt}{1.05}
\begin{tabular}{l|l|l|x{36}}
\multicolumn{1}{c|}{model}  & backbone  & \multicolumn{1}{c|}{modality}  & mAP \\
\shline
2-Stream~\cite{Sigurdsson2017} & VGG16 & RGB + flow & 18.6 \\
2-Stream +LSTM~\cite{Sigurdsson2017} & VGG16 & RGB + flow & 17.8\\
Asyn-TF~\cite{Sigurdsson2017} & VGG16  & RGB + flow & 22.4  \\
MultiScale TRN~\cite{Zhou2017} & Inception & RGB & 25.2  \\
I3D~\cite{Carreira2017} & Inception & RGB & 32.9  \\
I3D~\cite{xiaolongwang2017nonlocal}   & ResNet-101 & RGB & 35.5   \\
NL I3D~\cite{xiaolongwang2017nonlocal}  & ResNet-101  & RGB & {37.5}  \\
\hline
NL I3D + GCN   & ResNet-50 & RGB & 37.5  \\
I3D + GCN   & ResNet-101 & RGB & 39.1  \\
NL I3D + GCN   & ResNet-101 & RGB & \bd{39.7}  \\
\end{tabular}
\vspace{0.5em}
\caption{Classification mAP (\%) in the \textbf{Charades} dataset \cite{Sigurdsson2016}. NL is short for Non-Local.}
\label{tab:charades}
\vspace{-0.3in}
\end{table}

\paragraph{Training with a larger backbone.} Besides the ResNet-50 backbone architecture, we also verify our method on a much larger backbone model which is applied in~\cite{xiaolongwang2017nonlocal}. This backbone is larger than our baseline in 3 aspects: (i) instead of using ResNet-50, this backbone is based on the ResNet-101 architecture; (ii) instead of using $224 \times 224$ spatial  inputs, this backbone takes in $288 \times 288$ images; (iii)  instead of sampling 32 frames with 6fps, this backbone performs sampling more densely by using 128 frames with 24fps as inputs. Note that the temporal output dimension of both our baseline model and this ResNet-101 backbone are still the same ($16$ dimensions). With all the modifications on the backbone architecture, the FLOPs are 3 times as many as our ResNet-50 baseline model. 

We show the results together with all the state-of-the-art methods in Table~\ref{tab:charades}. The Non-local Net~\cite{xiaolongwang2017nonlocal} with ResNet-101 backbone achieves the mAP of $37.5\%$. We can actually obtain the same performance with our method by using a much smaller ResNet-50 backbone (with around $1/3$ FLOPs). By applying our method with the ResNet-101 backbone, our method (I3D+GCN) can still give $3.6\%$ improvements and reaches  $39.1\%$. This is another evidence showing that our method is modeling very different things from just increasing the spatial inputs and the depth of the ConvNets. By combining the non-local operation together with our approach, we obtain the final performance of $39.7\%$.

\vspace{-0.1in}
\subsection{Experiments on Something-Something}
\vspace{-0.05in}

In the Something-Something dataset, there are 86K training videos, around 12K validation videos and 11K testing videos. Each video has the duration ranging from 3 seconds to 6 seconds. The total number of classes is 174. 

The data in the Something-Something dataset is very different from the Charades dataset. In the Charades dataset, most of the actions are performed by agents in a  cluttered indoor scenes. However, in the Something-Something dataset, all videos are object centric and there is usually only one or two hands interacting with the center objects. The background in the Something-Something dataset is also very clean in most cases.

\begin{table}[t]
\centering
\small
\tablestyle{6pt}{1.05}
\begin{tabular}{l|l|x{28}x{28}|x{28}}
\multicolumn{1}{c|}{~}  & ~  & \multicolumn{2}{c|}{\emph{val}}  & \emph{test} \\
\multicolumn{1}{c|}{model}  & backbone  & top-1 & top-5  & top-1 \\
\shline
C3D~\cite{20bnsthsth} & C3D\cite{Tran2015} & - & - & 27.2  \\
MultiScale TRN~\cite{Zhou2017} & Inception & 34.4 & 63.2 & 33.6  \\
\hline
I3D  & ResNet-50 & 41.6 & 72.2 & -  \\
I3D + Spatial-Temporal GCN  & ResNet-50 & 42.8 & 74.7 & -  \\
I3D + Similarity GCN  & ResNet-50 & 42.7 & 74.6 & -  \\
I3D + Joint GCN  & ResNet-50 & 43.3 & 75.1 & -  \\
\hline
NL I3D & ResNet-50 & 44.4 & 76.0 & -  \\
NL I3D + Joint GCN  & ResNet-50 & 46.1 & 76.8 & 45.0  \\
\end{tabular}
\vspace{0.5em}
\caption{Classification accuracy (\%) in the \textbf{Something-Something} dataset \cite{20bnsthsth}. NL is short for Non-Local.}
\vspace{-0.2in}
\label{tab:20bn}
\end{table}

We report our results in Table~\ref{tab:20bn}. The evaluations are performed on both validation set and testing set. The baseline I3D approach achieves $41.6\%$ in top-1 accuracy and $72.2\%$ in top-5 accuracy. By applying our method with the I3D backbone (I3D + Joint GCN), we achieve $1.7\%$ improvements in the top-1 accuracy. We observe that the improvement of top-1 accuracy here is not as huge as the gains we have in the Charades dataset. The reason is mainly because the videos are already well calibrated with objects in the center of the frames. But interestingly, we still have a relative larger boost $2.9\%$ on the top-5 metric compared to the top-1 metric. We have also studied the performance of each sub-graph. If we only use spatial-temporal graph we obtain $42.8\%$ top-1 accuracy. With only similarity graph, we obtain $42.7\%$ accuracy.

We have also combined our method with the Non-local Net. As shown in Table~\ref{tab:20bn}, the Non-local I3D method achieves $44.4\%$ in top-1 accuracy. By combining our approach with the Non-local Net, we achieve another $1.7\%$ gain in top-1 accuracy, which leads to the state-of-the-art results $46.1\%$. We also test our final model on the test set by submitting to the official website. By using a single RGB model, we achieve the best result $45.0\%$ in the leaderboard.

\vspace{-0.1in}
\section{Conclusions}
\vspace{-0.1in}

We propose a novel graph based network to model the long range relationships in videos for action recognition. Large performance gain over state-of-the-art demonstrate the effectiveness of our framework. But more importantly, our error analysis shows how our model is doing better in capturing the object interactions, pose changes and actions in a sequence. This shows that our model has a large potential in not only video classification, but also variant tasks including detection and tracking in videos. 

\vspace{0.1in}

{\noindent {\bf Acknowledgement}: This work was supported by ONR MURI N000141612007, Sloan Fellowship, Okawa Fellowship to AG and Facebook Fellowship, NVIDIA Fellowship, Baidu Scholarship to XW. We would also like to thank Xinlei Chen, Gunnar Sigurdsson, Yin Li, Ross Girshick and Kaiming He for many helpful discussions.}

\clearpage

\bibliographystyle{splncs}
\bibliography{gcn}
\end{document}